%% file: main.tex
\documentclass{vgtc}                          

\pdfoutput=1

\ifpdf
  \pdfoutput=1\relax                   
  \usepackage{graphicx}                
  \DeclareGraphicsExtensions{.pdf,.png,.jpg,.jpeg} 
\else
  \usepackage{graphicx}                
  \DeclareGraphicsExtensions{.eps}     
\fi%

\graphicspath{{figures/}{pictures/}{images/}{./}} 

\usepackage{microtype}                 
\PassOptionsToPackage{warn}{textcomp}  
\usepackage{textcomp}                  
\usepackage{mathptmx}                  
\usepackage{times}                     
\usepackage{cite}                      
\usepackage{tabu}                      
\usepackage{booktabs}                  
\usepackage{color}
\newcommand{\squeezeup}{\vspace{-2.5mm}}

\onlineid{108}

\vgtccategory{Research}

\vgtcinsertpkg

\title{CNNComparator: Comparative Analytics of\\Convolutional Neural Networks}

\usepackage{authblk}
\author[1]{Haipeng Zeng\thanks{e-mail:tsanghaipeng@gmail.com}}
\author[1]{Hammad Haleem\thanks{e-mail:hammadhaleemhk@gmail.com}}
\author[1]{Xavier Plantaz\thanks{e-mail:xpfplantaz@connect.ust.hk}}
\author[2]{Nan Cao\thanks{e-mail:nan.cao@tongji.edu.cn}}
\author[1]{Huamin Qu\thanks{e-mail:huamin@cse.ust.hk}}
\affil[1]{The Hong Kong University of Science and Technology, Hong Kong}
\affil[2]{The Tongji University, China}

\abstract{Convolutional neural networks (CNNs) are widely used in many image recognition tasks due to their extraordinary performance. However, training a good CNN model can still be a challenging task. In a training process, a CNN model typically learns a large number of parameters over time, which usually results in different performance. Often, it is difficult to explore the relationships between the learned parameters and the model performance due to a large number of parameters and different random initializations. In this paper, we present a visual analytics approach to compare two different snapshots of a trained CNN model taken after different numbers of epochs, so as to provide some insight into the design or the training of a better CNN model. Our system compares snapshots by exploring the differences in operation parameters and the corresponding blob data at different levels. A case study has been conducted to demonstrate the effectiveness of our system.

} 

\CCScatlist{ 
  \CCScat{H.1.2}{Information Systems}%
{Models and Principles}{User/Machine Systems};
}

\teaser{
\centering
   \includegraphics[width=1\linewidth]{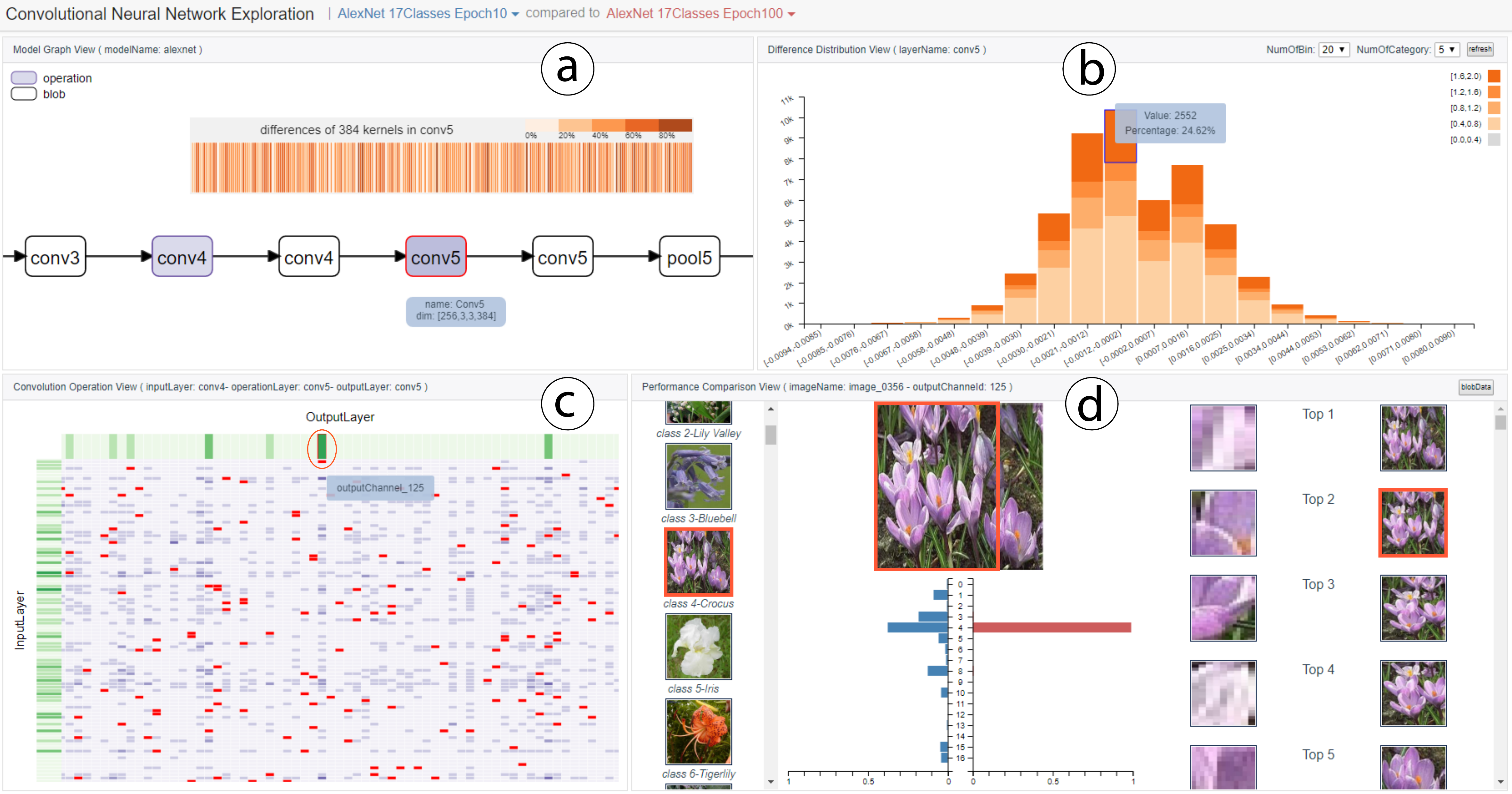}
   \squeezeup
   \caption{A visual analytics system for comparing two different snapshots of the AlexNet model after the 10th and 100th epochs. The network architecture view (a) shows the architecture of the Alexnet. The difference distribution view (b) shows the distribution of the parameter differences in a selected layer. The convolutional operation view (c) presents a selected convolutional operation as a 2D matrix to facilitate comparison. The performance comparison view (d) provides a side-by-side comparison of the model performance and image patches of top activation values.}
   \label{fig:systemOverview}
}

\begin{document}


\firstsection{Introduction}

\maketitle

\input{100-introduction}

\section{Related Work}
\input{200-relatedWork}

\section{Task Analysis}
\input{300-taskAnalysis}

\section{System Overview and Visual Design}
\input{400-systemOverview}

\section{Evaluation}
\input{500-evaluation}

\section{Discussion And Conclusion}
\input{600-conclusion}

\acknowledgments{
The authors would like to thank the anonymous reviewers for
their valuable comments. This project is partly supported by grant UIT/138 and  the National Basic Research Program of China (973 Program) under Grant No. 2014CB340304.}

\clearpage

\bibliographystyle{abbrv}

\bibliography{references}
\end{document}

%% file: 100-introduction.tex
Convolutional neural networks (CNNs) are widely used in many image recognition tasks, such as image classification~\cite{he2016deep, krizhevsky2012imagenet}, object detection~\cite{ren2015faster, liu2016ssd}, and video classification~\cite{karpathy2014large, zeiler2014visualizing}, due to their significant advantages over traditional machine learning methods. Despite such great success on these tasks, CNNs are still treated like black boxes due to their numerous parameters and unclear working mechanisms~\cite{bengio2013representation}. Thus, designing and training a good CNN model usually involves a substantial amount of trial and error.

Many studies have been conducted to comprehend the inner working mechanisms of deep neural networks, and visualization is often used due to its intuitiveness~\cite{liu2017towards, yosinski2015understanding, abadi2016tensorflow}. Most existing visualization works on CNNs mainly focus on what the neurons have learned, relationships in the network and the network structure, as well as the training information. However, existing CNN visualization techniques usually lack the ability to systematically explore and compare the differences in parameters/weights of two model snapshots. The performance of a model usually improves over time during the training process, but users normally can only obtain the training status from accuracy and loss information, so it is hard for them to know what happens to the parameters of the network and how they affect the performance of the CNN model. Thus, it would be helpful to get some insight into how model parameters evolve from a state with low accuracy to a state with high accuracy. For example, if we could visualize the update trend of parameters better, we could perhaps accelerate the training process and improve a model's performance.

In practice, training a CNN model involves learning the parameters for the model from a training dataset. The performance of the model should be highly related to these learned parameters. However, there are two main challenges in exploring the relationships between model parameters and performance: scalability and interpretability. For example, there may be millions of parameters in a model, which makes it hard to find some important parameters. Also, it is hard to interpret parameters' effects on the model's performance by directly exploring the parameters themselves. Thus, considering the large size of a typical neural network, we compare model snapshots by adopting a top-down analytical visualization method with different levels of detail, i.e., the model, layer, channel and neuron levels. For better interpretability, we explore the differences between two model snapshots from two aspects: the differences in operation parameters and the differences in the corresponding blob data given different image inputs. Since different training processes can sometimes yield significantly different results due to the differences introduced by random initialization, interpretation of different training process is not feasible. Thus, in this paper, we only focus on comparing the differences between two model snapshots in one training process after two different epochs.

To be specific, we present a visualization system called CNNComparator to compare two snapshots of a CNN model built for image classification. Our system contains four views. The network architecture view (Fig.~\ref{fig:systemOverview}a) provides an overview of the network structure and the differences among the layers. The difference distribution view (Fig.~\ref{fig:systemOverview}b) shows the distribution of the parameter differences in one selected layer. The convolutional operation view (Fig.~\ref{fig:systemOverview}c) simulates the entire convolution operation process, thereby showing the differences in the parameters and activation in detail. The performance comparison view (Fig.~\ref{fig:systemOverview}d) shows a comparison of the model performance on a given image and a comparison of image patches on a selected channel. The major contributions of this paper can be summarized as follows:
\begin{itemize}
\setlength\itemsep{0em}
    \item A visual analytics approach is proposed to enable users to explore and compare CNN operation parameters and blob activation at different levels.
    \item Several comparison designs with good scalability are applied to describe and compare two CNN model snapshots.
\end{itemize}

%
%
%
%
%
%
%
%
%
%
%
%
%
%
%
%
%

%% file: 200-relatedWork.tex
In this section, we summarize some relevant visualization work on CNNs and some approaches to visual comparison of data.

\subsection{Visualization on Convolutional Neural Networks}
Many researchers have focused on using visualization to better understand CNN models, which can be classified into three categories, namely, feature, relationship, and process visualization.

Most research has been conducted in the field of feature visualization, which aims to visualize features learned by neurons in a CNN. Existing work can be mainly divided into two categories, namely, code inversion~\cite{dosovitskiy2016inverting, mahendran2015understanding, zeiler2014visualizing} and activation maximization~\cite{erhan2009visualizing, simonyan2013deep, girshick2014rich, nguyen2016multifaceted, yosinski2015understanding}. The code inversion method aims to synthesize an image starting from the encoded image representation. Mahendran et al.~\cite{mahendran2015understanding} used a gradient descent optimization to invert representations. Activation maximization is another method that aims to find an image that maximally activates a neuron of interest, thereby revealing the features that the neuron has detected. However, these methods reveal only the features learned by neurons and fail to explore the structure of the network and the learned parameters of the network. Thus, in this paper we focus on analyzing the parameters of the network.

Apart from revealing what the neurons have learned in a CNN, researchers have also paid constant attention to relationships in a CNN model, i.e., relationships between representations and relationships between neurons. Rauber et al.~\cite{rauber2017visualizing} used the t-SNE dimension reduction technique and provided a detailed analysis of the projection, which confirms the known and reveals the unknown.  Liu et al.~\cite{liu2017towards} proposed CNNVis, which formulates a CNN as a directed acyclic graph and proposed hybrid visualization to reveal features learned by neurons. Their focus was on comparing how the depth and width of the network affect the training result, while our focus is on visually comparing the parameters of two CNN snapshots.

Some researchers are interested in the entire process of deep learning models and have proposed interactive systems~\cite{karpathy2014convnetjs, smilkovdirect, harley2015interactive, abadi2016tensorflow, deeplearning4j, Luke2016digits} to visualize neural network structures and training information; these systems can facilitate the learning and training of deep learning models. To the best of our knowledge, TensorBoard~\cite{abadi2016tensorflow} is perhaps the most famous and best visualization tool available to deep learning researchers. It visualizes neural networks as a computational graph, where users can check the status of the trained model and modify configurations. However, it shows the statistical summary and weight distribution information of the network directly without any further analysis. Thus, in this paper, we focus on comparative analytics of the parameters, which is more problem-driven.

\subsection{Visual Comparison Approaches}
Visually comparing two models with different parameters falls into the general topic of comparative visualization, especially in the area of weighted graph comparison and cube comparison.

Three major categories of comparative visualization approaches are summarized in previous work by Gleicher et al.~\cite{gleicher2011visual}; these categories are juxtaposition (i.e., side by side), superposition (i.e., overlay), and explicit encoding. Juxtaposition refers to placing objects separately and performing a side-by-side comparison, while superposition refers to putting multiple objects in the same coordinate system and then overlaying them to show the differences. Explicit encoding involves directly and visually encoding the differences between two objects. In one way, the entire CNN can be regarded as a graph, especially a weighted graph. Various comparison techniques that combine different visualization forms, such as a node-link diagram and an adjacency matrix, have been proposed~\cite{alper2013weighted, andrews2009visual}. In another way, a convolutional operation is in the form of a cube. Bach et al.~\cite{bach2016descriptive} presented a comprehensive and descriptive framework for temporal data visualization based on generalized space-time cubes. However, these methods are not ideal for CNN model comparison and cannot be easily adapted to compare the parameters of a CNN model due to the special structure of a CNN and numerous parameters. Borrowing ideas from existing work, we mainly use juxtaposition and explicit encoding to compare the data from a CNN model in this paper.

%% file: 300-taskAnalysis.tex
To better compare two different CNN model snapshots, we identified the following main questions:
\begin{enumerate}
\setlength\itemsep{0em}
\item [\textbf{Q1}] How different are two CNN model snapshots? How can we quantify differences between two model snapshots? 
\item [\textbf{Q2}] Where are these differences? Are there any patterns in these differences? 
\item [\textbf{Q3}] How can we find and highlight any correlation between the model performance and parameter changes?
\end{enumerate}

On the basis of the questions listed above, we identified the following analytic tasks:
\begin{enumerate}
\setlength\itemsep{0em}
\item [\textbf{T1}] \textbf{Global Exploration} Users should be able to intuitively observe the differences from an overview and quickly see major differences between two snapshots, such as the most different layer and differences in the model's performance (Q1).

\item [\textbf{T2}] \textbf{Detail Exploration} Once users select a layer to explore, users should be able to easily grasp what is different about this layer. A quick look should enable them to see the distribution and the parameters that change most (Q1).

\item [\textbf{T3}] \textbf{Insight Exploration} To gain further insight into differences, users should be able to easily locate differences, such as the position(s) where most weights change and the position(s) where most channels are activated (Q1 and Q2). 

\item [\textbf{T4}] \textbf{Correlation Exploration} On the basis of correlations between two model snapshots on a specific layer, users should be able to identify which features remain most relevant in both snapshots and observe the different ways a layer becomes activated when given the same input image (Q2 and Q3). 
\end{enumerate}


%% file: 400-systemOverview.tex
\begin{figure}[!htb]
   \centering
   \includegraphics[width=1\columnwidth]{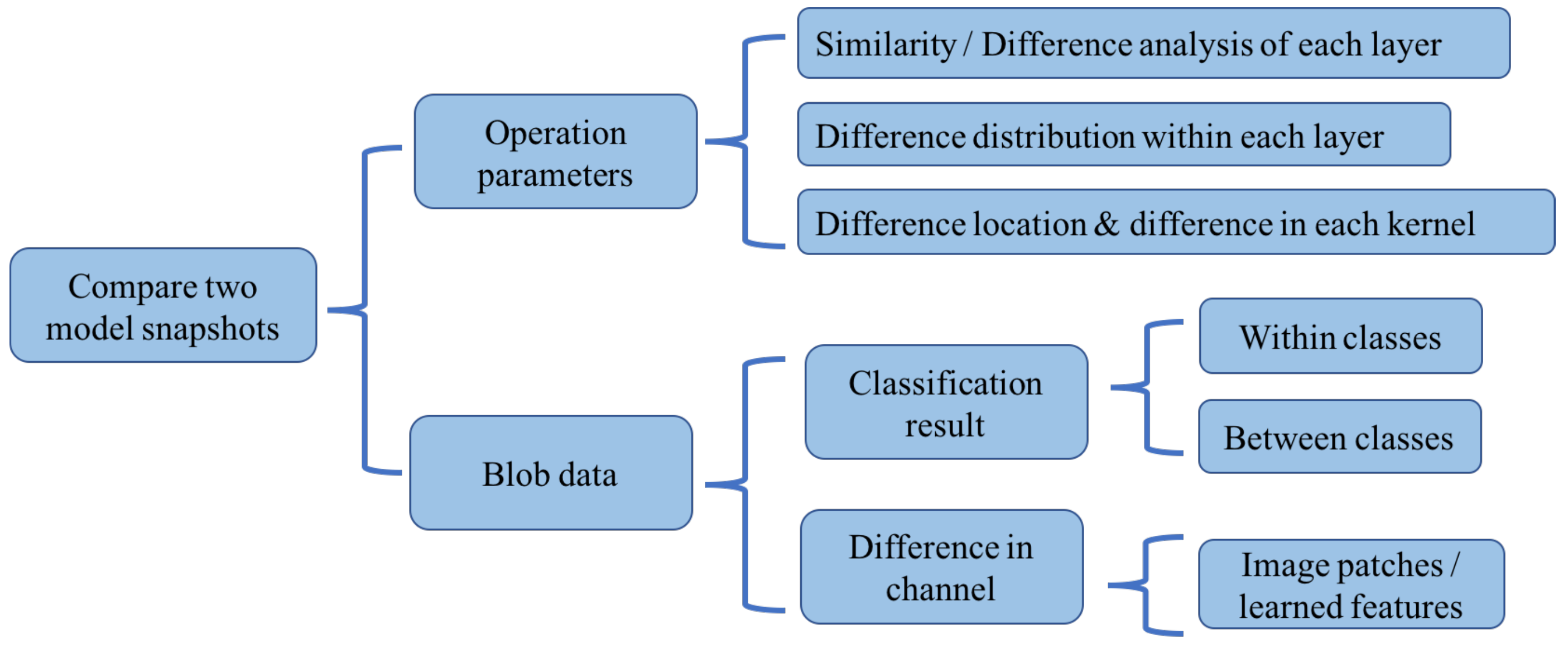}
   \caption{How our system compares two model snapshots. It explores the differences in operation parameters and the corresponding blob data.}
   \label{fig:comparisonStrategy}
\end{figure}

We designed our entire system (Fig.~\ref{fig:systemOverview}) based on the principle of ``overview first, zoom and filter, then details on demand''~\cite{shneiderman1996eyes}. As shown in Fig.~\ref{fig:comparisonStrategy}, we compare two model snapshots from two aspects: directly showing the differences in operation parameters and analyzing the differences in corresponding blob data when given an input image. 

\subsection{Network Architecture View}
This view (Fig.~\ref{fig:systemOverview}a) shows the network architecture of the model, which provides an overview of what the model looks like. We simply use Euclidean distance to show the differences in parameters of each operation layer. Color is used to encode the difference (a darker shade indicates greater difference), thereby assisting users in identifying a layer for further exploration (T1). When users hover on one interesting layer, the corresponding information will be shown up, i.e., the name, the shape and overall difference in kernels of this layer. The upper part of Fig.~\ref{fig:systemOverview}a can give a hint on how much difference the kernels have in the interesting layer. The darker color means more difference. Ideally, users can click on the layer that has changed most/least to obtain a comprehensive comparison in the next three views.

\subsection{Difference Distribution View}
We use a histogram and a stacked chart in this view (T2). As actual parameter weights are quite small and near zero, and users usually focus more on relative change rather than absolute change, this view is designed to highlight relative changes. First, we create bins based on the absolute change in weights. Then, for each bin, we quantify relative change into several change levels, which are indicated by different colors. We adopt relative percent difference~\cite{tornqvist1985should}, because the normal relative change method is not suitable when the denominator is close to zero. 

\begin{equation}
d(x,y)=\frac{|x-y|}{(|x|+|y|)/2}=2 \frac{|x-y|}{|x|+|y|}\in [0,2]
\end{equation}
where x, y denotes two variables whose relative change we aim to measure. d(x, y) denotes the unsigned relative change whose value lies between 0 and 2. A large value corresponds to considerable change between two variables.

In the difference distribution view (Fig.~\ref{fig:systemOverview}b), the size of the rectangle encodes the number of weights in the corresponding category. The color encodes the relative change in different levels. A darker color indicates more changes, whereas a lighter color indicates few changes. Users can filter different change levels if some of them are dominant. If users are more interested in one category, they can click the  corresponding rectangle, and the corresponding weights will be highlighted in the convolutional operation view, thereby allowing users to know the location of these weights (T3).

\begin{figure}[!htb]
   \centering
   \includegraphics[width=0.9\columnwidth]{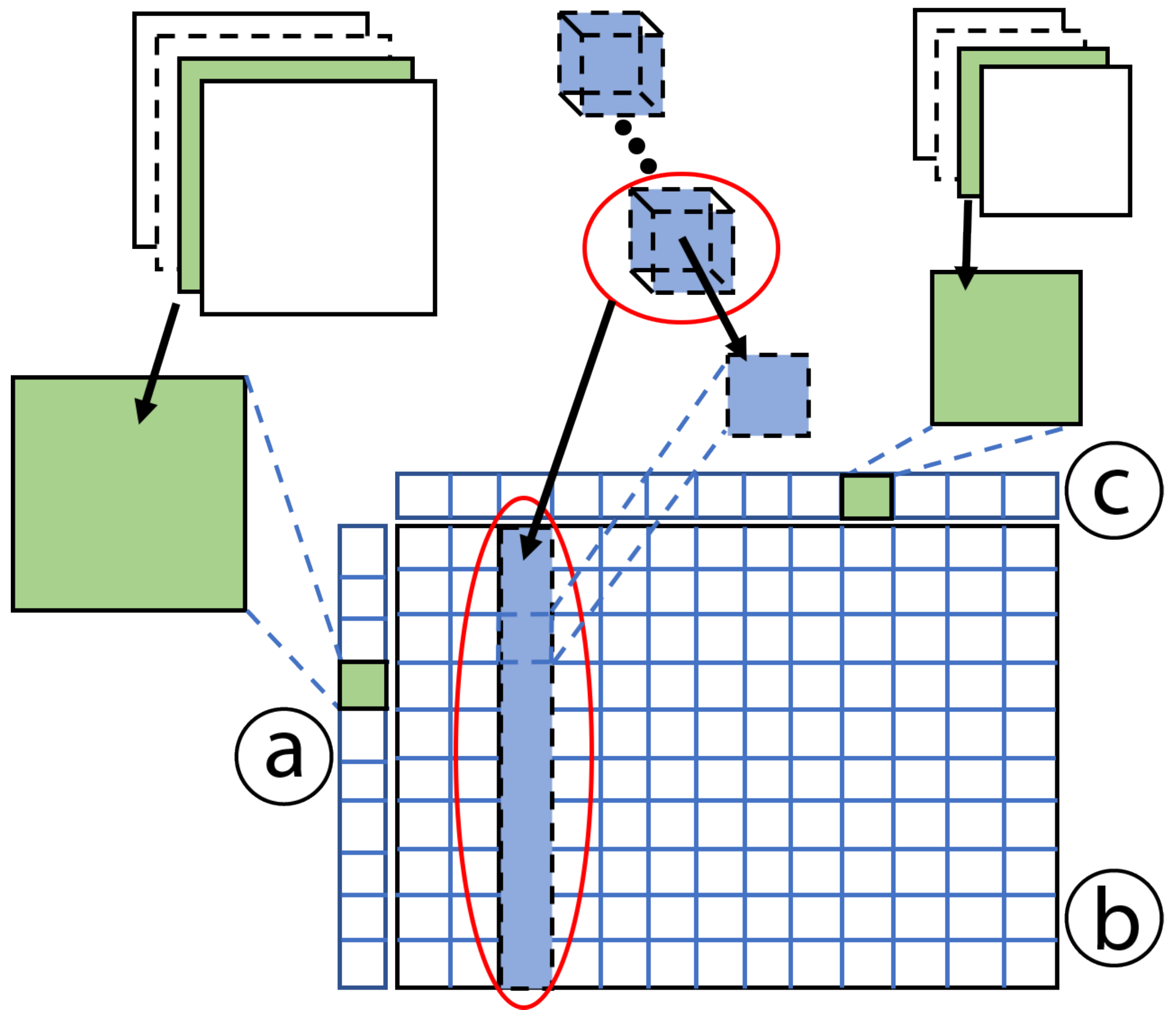}
   \caption{A 2D convolutional operation pixel map. In the input layer (a), each rectangle (green) represents one channel map; in the convolutional operation layer (b), each column (blue) in the matrix represents a kernel, and each rectangle in the column represents one channel kernel map; in the output layer (c), each rectangle (green) represents one channel map.}
   \label{fig:convolutionOperation}
\end{figure}

\subsection{Convolutional Operation View}
To facilitate better understanding and fast comparison of convolutional operation, we present a 2D convolutional operation pixel map (Fig.~\ref{fig:systemOverview}c), which transforms 4D convolutional convolution operation kernels into a 2D matrix pixel view. Fig.~\ref{fig:convolutionOperation} illustrates our design. Three components exist in this view, namely, input, output, and operation layers. The operation layer corresponds to the selected convolutional layer, and the input and output layers correspond to the previous layer and next layer of the selected convolutional layer, respectively. Each rectangle in the input and output layers indicates one channel map of the corresponding layer in the network. Each column of the operation layer represents one kernel of the convolutional operation, and each rectangle in the column shows different channels of the kernel. We simply use color to encode the difference between two model snapshots in this view, which can help users identify the differences in different channels (T3). 

Typically, a tremendous amount of parameters exist, thereby giving rise to a scalability issue. To solve this problem, we added some user interactions, such as zoom in/out to explore a part of interest, hover on to show the position information, and click to pop up a corresponding kernel matrix. When users click a rectangle in the input or the output layer, the corresponding image patches are shown in the performance comparison view (T4).

\subsection{Performance Comparison View}
In this view (Fig.~\ref{fig:systemOverview}d), we mainly adopt side-by-side comparison on the performance of two model snapshots (T1). Users can select an input image. Then, the classification result is shown side-by-side in bar charts, thereby making the observation of the distribution of probability for each class easier for users to read. To allow users to explore more information, we further employ a similar method used in~\cite{girshick2014rich} to compute learned features of neurons. We use selective search~\cite{uijlings2013selective} to crop some image patches and then rank them by activation value on the selected channel (T4). When users hover on the image patches, the corresponding positions will be highlighted in the original image.

%% file: 500-evaluation.tex
CNNComparator is a web-based system to explore how different network parameters and activation values are between two CNN model snapshots. In our evaluation, we train the AlexNet~\cite{krizhevsky2012imagenet} model based on the TFLearn framework on the 17-category flower dataset~\cite{nilsback2006visual}, where each category has 80 images. After running the model for 100 epochs, we obtained $97.2\%$ accuracy on the training set and $72.79\%$ accuracy on the validation set. 

\subsection{Case Study}
\begin{figure}[!htb]
   \centering
   \includegraphics[width=1\columnwidth]{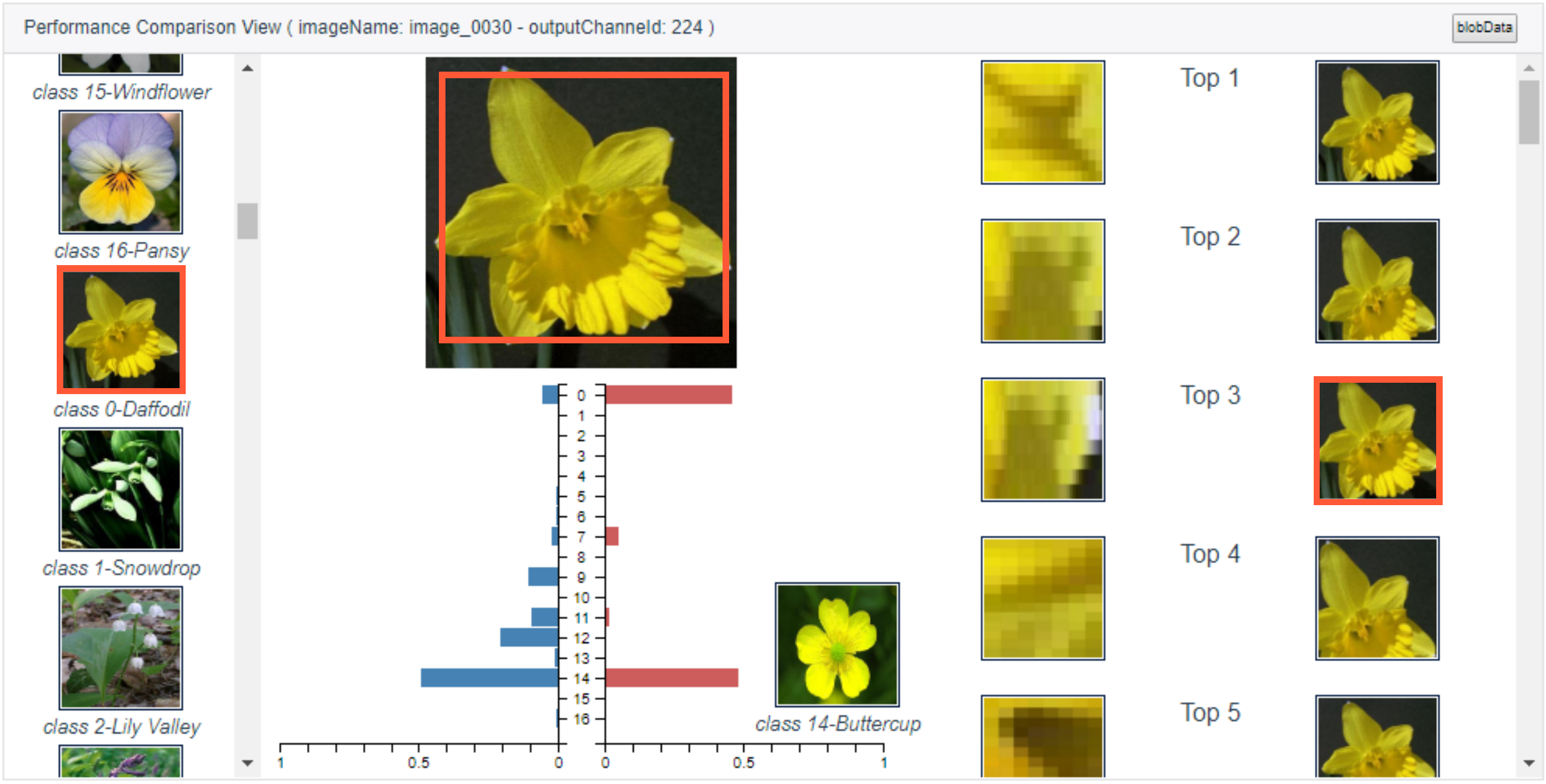}
   \caption{A comparison result of snapshots of a trained CNN model taken after 10 and 100 epochs. Although the model after 100 epochs outperforms the model after 10 epochs; it mistakenly classifies an image of a daffodil as a buttercup, since the feature extraction of snapshot 100 mainly focuses on yellow.}
   \label{fig:case}
\end{figure}

We did a case study to show the effectiveness of our system. We chose two snapshots in the same training process for comparison, namely, epoch 10 and epoch 100. Usually, for a model designer, the crucial feedback from the model is related to the accuracy and loss of a snapshot. By using our system, users can explore the finer aspects of the model. As shown in Fig.~\ref{fig:systemOverview}d, we can easily see that the model after epoch 100 outperforms the model after epoch 10 considerably after an input image is selected. The network architecture view (Fig.~\ref{fig:systemOverview}a) shows a considerable difference in the conv5 layer. Thus, we clicked it and then the difference distribution view showed the distribution of differences between the two snapshots. As we can see, minor changes in the parameters between two snapshots are dominant. To determine which part(s) changed significantly, we filtered out small value changes by clicking the corresponding legend (the color changes to gray) and then clicking the rectangle we were interested in; the corresponding locations were highlighted in the convolutional operation view (Fig.~\ref{fig:systemOverview}c). At first glance, they were quite randomly distributed; but upon careful observation, we could see that some neurons were highly activated. As for blob data that corresponds to the selected input image, some channels were darker than others, thereby indicating more differences were in these channels between two snapshots. Once we clicked on a channel, we could see the top image patches shown in Fig.~\ref{fig:systemOverview}d. Clearly, at epoch 100, the model is more likely to capture more information (detailed image patches), whereas at epoch 10, the model is more focused on some abstract image patches. In the snapshot after epoch 10, the image had a higher probability of being misclassified, mainly because features are not detailed enough. Fig.~\ref{fig:case} demonstrates the same pattern. However, even in the snapshot after epoch 100, the model was likely to classify the image to wrong class (class 14). After careful exploration, we found that this was because yellow is the dominant color in class 14, and this snapshot mainly captures the yellow feature.

%% file: 600-conclusion.tex
In this paper, we introduced an interactive visualization system called CNNComparator, which includes four linked views that enable users to explore the evolution of CNN parameters over one training process and explore generated learned features. Our system provides a top–down approach for a comparative analysis of two model snapshots. The evaluation part demonstrated the usefulness of CNNComparator in exploring the differences between CNN snapshots. Although this study focuses on AlexNet as a primary candidate for evaluation, this approach could be easily extended and applied to other neural network architectures (e.g., VGG, GoogleNet) since they have similar convolutional operation structures.

However, there are some limitations of this work. First, given the numerous neuron parameters, scalability is still a major challenge in visualizing and comparing two model snapshots. Visualizing all neuron parameters directly will easily cause severe visual clutter. Also, we observed that the locations of changes are quite random. So, more interactions should be added to filter out useless information by leveraging domain knowledge. Second, we applied some simple metrics in this paper, such as Euclidean distance. In order to better measure the differences, some other alternative metrics can be explored. Third, currently we focused on only two snapshots in one training process, and we did not compare snapshots from different training processes with the same or different architectures. It might be more useful to compare two different models with different architectures or hyper-parameters. However, it is very challenging to directly conduct visual comparison on them due to different model structures and very different random initializations of the parameters inside networks. One of the major issues we need to consider is how to interpret the comparison results affected by different random initializations. Therefore, in our paper, we start to address an easier problem by comparing two model snapshots in one training process. Also, it might be more natural to show temporal trends of how models change over epochs, rather than picking two numbers. In the future, we will try to explore more complicated cases and extend this approach to different training processes and different model structures, as well as show temporal trends of how models change over epochs.

